\documentclass[10pt, a4paper]{article}

\usepackage[final]{lrec2026}
\usepackage{amsmath}
\usepackage{booktabs}
\usepackage[most]{tcolorbox}
\newcommand{\textdetox}{\textsc{TextDetoxEval}}

\title{Evaluating Text Style Transfer: \\ A Nine-language Benchmark for Text Detoxification}

\name{Vitaly Protasov$^1$, Nikolay Babakov$^2$, Daryna Dementieva$^{3,4}$, Alexander Panchenko$^{5,1}$} 

\address{$^1$AIRI \quad $^2$Universidade de Santiago de Compostela \quad $^3$Technical University of Munich (TUM)\\ $^4$Munich Center for Machine Learning (MCML) \quad $^5$Skoltech \\
         \href{mailto:protasov@airi.net}{protasov@airi.net}, \href{mailto:nikolay.babakov@usc.es}{nikolay.babakov@usc.es}, \href{mailto:daryna.dementieva@tum.de}{daryna.dementieva@tum.de}, \href{mailto:a.panchenko@skol.tech}{a.panchenko@skol.tech}}

\abstract{
Despite notable advances in large language models (LLMs), reliable evaluation of text generation tasks such as text style transfer (TST) remains an open challenge. Existing research has shown that automatic metrics often correlate poorly with human judgments~\cite{Dementieva2024OverviewOT,DBLP:journals/corr/abs-2502-15022}, limiting our ability to assess model performance accurately. Furthermore, most prior work has focused primarily on English, while the evaluation of multilingual TST systems, particularly for text detoxification, remains largely underexplored.
In this paper, we present the first comprehensive multilingual benchmarking study of evaluation metrics for \textbf{text detoxification} evaluation across \textit{nine languages}: Arabic, Amharic, Chinese, English, German, Hindi, Russian, Spanish, and Ukrainian. 
Drawing inspiration from machine translation evaluation, we compare neural-based automatic metrics with LLM-as-a-judge approaches together with experiments on task-specific fine-tuned models. Our analysis reveals that the proposed metrics achieve significantly higher correlation with human judgments compared to baseline approaches. We also provide actionable insights and practical guidelines for building robust and reliable multilingual evaluation pipelines for text detoxification and related TST tasks.
 \\ \newline  
 }

\begin{document}

\maketitleabstract

\section{Introduction}

Evaluation of text generation tasks remains a long-standing challenge in natural language processing (NLP), as suitable metrics must capture both task-specific objectives and linguistic variation across languages. The difficulty of designing such metrics increases with the number of languages considered, since stylistic, morphological, and pragmatic differences demand diverse evaluation approaches. Over the past decade, the field has seen continuous evolution of evaluation methodologies: from character-based measures such as \textsc{chrF}~\cite{Popovic2015chrFCN}, to lexical and semantic metrics like ROUGE~\cite{Lin2004ROUGEAP}, METEOR~\cite{Banerjee2005METEORAA}, and BERTScore~\cite{Zhang2019BERTScoreET}, and more recently to neural-based models such as COMET~\cite{Rei2020COMETAN}. With the emergence of large language models (LLMs), new evaluation paradigms---particularly \textit{LLM-as-a-judge} setups---have demonstrated strong potential for approximating human judgments~\cite{Bavaresco2024LLMsIO}.

In this study, we focus on a specific text style transfer (TST) task: \textbf{text detoxification}. The goal of this task is to remove toxic or offensive content from text while preserving the original meaning and fluency. Beyond its research value, text detoxification is of practical importance for a wide range of real-world applications, including online moderation, dialogue systems, and social media content filtering and post-processing. For instance, consider a video streaming service which can offer a dynamic rewriting of abusive words enabling a child-friendly watching. Another important use-case is to implement guards against an eventual generation of abusive words  by a neural dialogue system for client support.

Despite its relevance, the task remains underexplored in both data and evaluation methodology. To date, only two publicly available datasets exist: \textdetox~\cite{Dementieva2024OverviewOT}, covering nine languages, and DialogueEvaluation-2022~\cite{Dementieva2022RUSSE2022}, which focuses solely on Russian. Moreover, evaluation practices in this area are often inconsistent, with most prior studies relying on generic metrics rather than task-specific or multilingual approaches.

To address these gaps, we perform a comprehensive experimental study of evaluation metrics for the text detoxification task across \textbf{nine languages}: Arabic, Amharic, Chinese, English, German, Hindi, Russian, Spanish, Ukrainian. Our goal is to analyze the behavior, limitations, and cross-lingual robustness of different metric families, including both automatic neural-based metrics and LLM-as-a-judge approaches. In addition, we perform fine-tuning experiments with one of open-source LLMs on annotated data. We aim to identify metrics that align more closely with human judgments and to provide guidance for future evaluation of detoxification and related TST systems.
The main contributions of this work are:

\begin{enumerate}
    \item We conduct an extensive multilingual evaluation study for the text detoxification task, covering all publicly available datasets for this task.
    \item We experiment with a diverse set of evaluation metrics, proposing several improved metric configurations tailored to text detoxification.
    \item We compare all approaches against existing automated and LLM-as-a-judge setups, highlighting their respective strengths and weaknesses across languages.
    \item We fine-tune open-source LLMs on annotated detoxification data and evaluate their suitability as automatic judges.
\end{enumerate}

To facilitate future research and reproducibility, we make our evaluation setup, code, and results publicly available.\footnote{\scriptsize{\href{https://github.com/textdetox/eval-of-detox-eval}{https://github.com/textdetox/eval-of-detox-eval}}} Details of all resources, corresponding links, and their licenses are presented in Appendix~\ref{sec:app_licenses}.

\section{Related Work}
\subsection{Automatic Evaluation for TST and Text Detoxification}
\label{sec:related_work1}
The task of text style transfer (TST) has been studied across a variety of domains. This includes sentiment transfer, such as converting between positive and negative reviews~\cite{li-etal-2018-delete}; formality transfer~\cite{rao-tetreault-2018-dear,briakou-etal-2021-ola}, which focuses on transforming informal texts into formal ones; and stylistic rewriting, exemplified by the Bible style transfer task~\cite{Carlson2018}, which leverages translations from different historical periods. Additionally, the biased-to-neutral Wikipedia corpus~\cite{pryzant2020automatically} makes use of editorial revisions to reduce bias. In the domain of text detoxification, the task has been addressed already with several solutions using both unsupervised approaches~\cite{nogueira-dos-santos-etal-2018-fighting,dale-etal-2021-text,hallinan-etal-2023-detoxifying} and supervised methods, supported by parallel data~\cite{logacheva-etal-2022-paradetox,sourabrata-etal-2023-text,smurfcat_at_pan}.
Across various domains, the evaluation of text style transfer systems has traditionally relied on three core criteria:
\begin{itemize}
    \setlength{\itemsep}{1pt}
    \setlength{\parskip}{1pt}
    \setlength{\parsep}{1pt}
    \item \textit{Style Accuracy} (\textbf{STA}): the proportion of outputs correctly classified in a new style by a style classifier.

    \item \textit{Content Preservation} (\textbf{SIM}): the extent to which the key semantic content from the original input is retained.

    \item \textit{Fluency} (\textbf{FL}): whether the generated output maintains natural fluency or, at minimum, does not degrade the fluency of the original.
\end{itemize}
Considerable efforts have been made to develop more robust evaluation metrics for text style transfer~\cite{DBLP:journals/corr/abs-2406-18403,DBLP:journals/corr/abs-2502-15022} and, in particular, for text detoxification~\cite{dementieva-etal-2023-exploring}, a universally accepted automatic evaluation framework that strongly correlates with human judgment has yet to be established. Thus, as a recent state-of-the-art evaluation setup, we adopt setup from \citet{Dementieva2024OverviewOT} where: (i) \textbf{STA} is defined as the probability assigned by a pre-trained XLM-R~\citep{Conneau2019UnsupervisedCR} toxicity classifier indicating that the output text belongs to the neutral (non-toxic) class;
(ii) \textbf{SIM} is calculated as the cosine similarity between LaBSE \citep{Feng2020LanguageagnosticBS} representations of toxic and detoxified texts;
(iii) \textbf{FL}: measured as a proxy using the ChrF \citep{Popovic2015chrFCN}, which compares the generated detoxified text to human-written references. 
These all three parameters are then combined into joint metric \textbf{J}:
\begin{equation} \label{eq:J}
\begin{aligned}
    \textbf{J} = \frac{1}{n}\sum\limits_{i=1}^{n}\textbf{STA}(y_i) \cdot \textbf{SIM}(x_i,y_i) \cdot \textbf{chrF}(x_i, y_i),
\end{aligned}
\end{equation}
where \textbf{STA}($y_i$), \textbf{SIM}($x_i,y_i$), \textbf{chrF}($x_i,y_i$) $\in [0, 1]$ for each text detoxification output $y_i$. Here $x_i$ means detoxified golden texts and $y_i$ is detoxified generated texts.

\paragraph{Limitations of Current Metrics}
\label{sec:metric_limitations}
Current evaluation metrics for detoxification are hindered by their limited and often superficial use of human reference texts. 
Thus, only \textit{fluency}, assessed via ChrF, explicitly leverages references. However, ChrF suffers from a core limitation: it evaluates surface-level n-gram overlap with the reference, ignoring the semantic relationship between the system output and the original toxic input. 
This reliance on lexical similarity renders the metric both \textit{overconstrained}---discouraging variation---and \textit{underconstrained}---failing to ensure semantic preservation. These issues underscore the need for evaluation approaches that more effectively integrate both source and reference relationships.

\subsection{LLMs as a Judge} The emergence of large language models (LLMs) has introduced a new paradigm in evaluation, where LLMs themselves are used as judges for NLP tasks~\cite{DBLP:journals/corr/abs-2412-05579}. This approach has been explored by \citet{DBLP:journals/corr/abs-2502-15022} across 20 NLP tasks, including text style transfer (TST). In the domain of abusive language, LLMs have been employed to assess the relevance and appropriateness of counter-speech responses to hate speech~\cite{jones-etal-2024-multi,bonaldi-etal-2024-nlp}. TST can also be framed as a paraphrasing task, where LLMs-as-judges have shown potential~\cite{lemesle-etal-2025-paraphrase}. While LLM-based evaluation is not without limitations, it offers a promising and adaptable solution---particularly for multilingual contexts.

\section{Evaluation Datasets}
\label{data_desc}

Our experiments rely on two available datasets for text detoxification task: \textdetox~\cite{Dementieva2024OverviewOT} and DialogueEvaluation-2022~\cite{Dementieva2022RUSSE2022}, with most of the focus on the first one due to its multilinguality.

\subsection{\textdetox}

\textdetox\footnote{\scriptsize{\href{https://hf.co/datasets/textdetox/detoxification_pairwise_style_evaluation}{https://hf.co/datasets/textdetox/detoxification\_pairwise\_\\style\_evaluation}}}$^,$\footnote{\scriptsize{\href{https://hf.co/datasets/textdetox/detoxification_pairwise_content_evaluation}{https://hf.co/datasets/textdetox/detoxification\_pairwise\_\\content\_evaluation}}} is a multilingual dataset released as part of the CLEF Text Detoxification Shared Task. It includes manual assessments of 20 detoxification systems across 9 languages: Amharic, Arabic, Chinese, English, German, Hindi, Russian, Spanish, and Ukrainian. The systems covered a range of modeling strategies, from unsupervised to fine-tuned and LLM-based prompting approaches~\cite{DBLP:conf/clef/PengHZYLLGCLT24,DBLP:conf/clef/LuoLW24,DBLP:conf/clef/Protasov24}.
For each language, 100 toxic sentences were selected, resulting in 900 inputs overall. Each of the 20 participant generated detoxified versions, producing \textbf{16,600} input–output pairs in total with corresponding human scores. Native speakers annotated the data on Toloka.ai\footnote{\scriptsize{\href{https://toloka.ai}{https://toloka.ai}}} according to three criteria:
\begin{itemize}
    \setlength{\itemsep}{0pt}
    \setlength{\parskip}{0pt}
    \setlength{\parsep}{0pt}
    \item \textbf{Fluency:} grammaticality and readability rated as \textit{yes}, \textit{partially}, or \textit{no}.
    \item \textbf{Content Similarity:} whether the detoxified text preserves the meaning of the original (binary judgment).
    \item \textbf{Style Transfer Accuracy:} which text is more toxic (\textit{original}, \textit{detoxified}, or \textit{neither}), randomized to avoid bias.
\end{itemize}
\subsection{DialogueEvaluation-2022}

DialogueEvaluation-2022\footnote{\scriptsize{\href{https://hf.co/datasets/textdetox/humaneval_textdetox_ru}{https://hf.co/datasets/textdetox/humaneval\_textdetox\_ru}}} is a Russian-language text detoxification dataset released as part of the first shared task on detoxification. The dataset contains toxic sentences collected from social media platforms. It includes outputs from \textbf{15 participating solutions}, each generating detoxified rewrites for a shared test set of \textbf{875 toxic sentences}.

Data collection followed a three-stage crowdsourcing procedure on Toloka. First, annotators rewrote toxic sentences into fluent, non-toxic paraphrases while preserving meaning. Second, independent annotators verified semantic equivalence between the original and rewritten texts. Third, additional annotators checked that toxicity had been successfully removed. Only paraphrases that passed both verification stages with at least 90\% agreement were retained.
Each text pair was further evaluated along three dimensions consistent with the \textdetox\: \textit{fluency}, \textit{content preservation}, and \textit{style transfer quality}. This dataset provides a valuable complementary resource for benchmarking Russian-language detoxification and cross-lingual metric evaluation.

Although DialogueEvaluation-2022 offers useful monolingual insights, we focus the main analysis of this paper on the multilingual \textdetox\ dataset to ensure consistency across languages and to maintain a unified evaluation setup. The detailed results for DialogueEvaluation-2022 are provided in Appendix \ref{sec:appendix_russe2022}.

\section{Evaluation Methodology}
\label{sec:eval_methodology}
In this section, we first aim to describe how evaluation in text detoxification task has been performed in previous works~\cite{Dementieva2022RUSSE2022, Dementieva2024OverviewOT}, and second, to propose ways to improve it by exploring alternative approaches or improving existing ones. Afterwards, we conduct experiments based on the described approaches to benchmark them and assess whether our proposed improvements are effective.
\subsection{Fluency}

As noted in Section~\ref{sec:metric_limitations}, current fluency evaluation in text detoxification typically relies on ChrF scores calculated between system outputs and human references. However, this approach ignores the original toxic input. As a result, it may favor outputs that look similar to the reference but do not fully preserve the meaning or context of the input sentence. In detoxification tasks, this becomes a problem because the goal is not only to make the text fluent and non-toxic, but also to preserve the meaning and context of the original text.

To address this limitation, we aim to improve fluency evaluation by considering more advanced neural-based models, namely \textbf{COMET}~\citep{Rei2020COMETAN} and \textbf{XCOMET}~\citep{Guerreiro2023xcometTM}. 
%In addition to ChrF, we also consider METEOR~\citep{Banerjee2005METEORAA}, which accounts for synonymy and stemming, as well as more advanced neural-based evaluation models, namely \textbf{COMET}~\citep{Rei2020COMETAN} and \textbf{XCOMET}~\citep{Guerreiro2023xcometTM}. 
Unlike traditional metrics that rely solely on lexical overlap with a reference, COMET-based models use pretrained encoders to model semantic relationships between the \textit{input}, \textit{system output}, and \textit{reference}. This triplet-based setup allows them to jointly assess whether the generated text maintains the original meaning and intent.
In machine translation, COMET models demonstrated significantly higher correlation with human judgments compared to $n$-gram metrics, as they better capture meaning, syntax, and fluency beyond token overlap.

In the context of text detoxification, we hypothesize that COMET models are suitable, as they should better capture the trade-off between fluency, content preservation, and toxicity reduction, which are key aspects that simpler lexical metrics often fail to account for. In our experiments, we consider 4 COMET variants:
\begin{enumerate}
    \setlength{\itemsep}{0pt}
    \setlength{\parskip}{0pt}
    \setlength{\parsep}{0pt}
    \item \textbf{Unbabel/wmt22-comet-da}~\cite{Rei2022COMET22U2}: A regression-based model trained on direct assessment data from WMT22, representing a standard reference-aware evaluation setting.
    \item \textbf{Unbabel/XCOMET-XL}~\cite{Guerreiro2023xcometTM}: A multilingual extension of COMET with a 3.5B-parameter encoder, improving correlation with human judgments across languages.
    \item \textbf{Unbabel/XCOMET-XXL}~\cite{Guerreiro2023xcometTM}: A 10.7B-parameter variant achieving state-of-the-art performance on WMT22 evaluation benchmarks.
    \item \textbf{myyycroft/XCOMET-lite}~\cite{larionov-etal-2024-xcomet}: A compressed and quantized model retaining over 95\% of XXL’s performance while reducing computational overhead by 60\%, enabling scalable multilingual evaluation.
\end{enumerate}
\subsection{Content similarity}

The content similarity score measures how well the generated text preserves the main semantic information of the original input. This metric penalizes outputs that lose important details or change the original meaning during the detoxification process. As noted in  Section~\ref{sec:metric_limitations}, previously content similarity was computed using cosine similarity between the embedding representations of source toxic texts and their detoxified versions. This approach completely ignores available reference texts, which creates two significant problems. First, metrics that only compare input to output perform poorly when the generated text requires substantial rewording—exactly what detoxification demands~\cite{shen-etal-2022-evaluation}. Second, without references, the metric cannot determine whether semantic content is genuinely preserved when lexical choices differ substantially from the input~\cite{shen-etal-2022-evaluation}.

To address these limitations, we propose an improved content similarity measurement metric that jointly considers both the input–output and output–reference relationships:

\begingroup
\setlength{\abovedisplayskip}{-7pt}
\setlength{\belowdisplayskip}{7pt}
\setlength{\abovedisplayshortskip}{0pt}
\setlength{\belowdisplayshortskip}{0pt}
\begin{equation} \label{eq:1}
\begin{aligned}
\text{c}_{\text{sim}} = w_{\text{i,g}} \cdot \cos_{\text{sim}}(v_{\text{i}}, v_{\text{g}}) 
+ w_{\text{g,r}} \cdot \cos_{\text{sim}}(v_{\text{g}}, v_{\text{r}})
\end{aligned}
\end{equation}
\endgroup

\noindent where $v_{\text{i}}$, $v_{\text{g}}$, and $v_{\text{r}}$ denote the embedding representations of the \textbf{input}, \textbf{generated}, and \textbf{reference} texts, respectively. The weights $w_{\text{i,g}}$ and $w_{\text{g,r}}$ control the contribution of input–output and output–reference similarity, with $w_{\text{i,g}} + w_{\text{g,r}} = 1$. This approach aims to balance two complementary aspects of content preservation: semantic consistency with source inputs and alignment with human-produced references, which together provide a more complete estimate of content preservation.

\subsection{Style transfer performance: toxicity}
\label{sec:sta_desc}
Toxicity measurement reflects how well a system transforms toxic inputs into neutral or polite outputs. In previous works, toxicity was measured using the probability of the generated text being classified as \textit{neutral} by a binary toxicity classifier. However, relying solely on this probability introduces several limitations: 
(i) high dependence on the biases and calibration of the specific classifier; 
(ii) toxicity of the original input is not taken into account, making it impossible to measure relative improvement; and 
(iii) reference texts, which represent human-level detoxification quality, are ignored entirely.

To provide a more robust assessment, we propose evaluating style transfer through comparative probability analysis across three text variants:
\begin{itemize}
    \setlength{\itemsep}{1pt}
    \setlength{\parskip}{1pt}
    \item Input (toxic) text: $P_{\text{neutral}}(t_{\text{i}})$,
    \item Generated text: $P_{\text{neutral}}(t_{\text{g}})$,
    \item Reference (neutral) text: $P_{\text{neutral}}(t_{\text{r}})$.
\end{itemize}

This setup enables the evaluation to capture the relative change in toxicity rather than relying on an absolute classifier score. In other words, instead of only checking whether an output is predicted as neutral, we measure how well it has improved compared to its toxic input, and whether it approaches the quality of a human-written neutral reference. Such relative comparisons are less sensitive to classifier calibration and better reflect the underlying goal of detoxification — to reduce toxicity while preserving semantic content.

To ensure consistent behavior across diverse systems, we introduce two stabilization rules:
\begin{enumerate}
    \setlength{\itemsep}{2pt}
    \setlength{\parskip}{0pt}
    \setlength{\parsep}{0pt}
    \item \textbf{Penalization:} if the generated text is more toxic than the input 
    ($P_{\text{neutral}}(t_{\text{g}}) < P_{\text{neutral}}(t_{\text{i}})$), 
    the score is set to zero: $Score(t_{\text{g}}) = 0$;
    \item \textbf{Rewarding:} if the generated text achieves neutrality equal to or greater than the reference 
    ($P_{\text{neutral}}(t_{\text{g}}) \ge P_{\text{neutral}}(t_{\text{r}})$), 
    the score is set to one: $Score(t_{\text{g}}) = 1$.
\end{enumerate}
This triplet-based design makes the toxicity metric more robust as it mirrors the logic used in COMET and content similarity metrics, where relationships between input, output, and reference are jointly modeled. As a result, the metric better reflects meaningful improvements in detoxification quality and mitigates dependence on specific classifiers.

\subsection{LLMs as judges}
\label{sec:sec4-llmjudge}
To complement proposed automatic metrics and assess their consistency with human evaluations, we additionally employ LLMs as automatic evaluators to compare how well different metric-based families (lexical, neural, and LLM-based) align with human annotations in the context of text detoxification.

\subsection{Fine-tuning of LLMs}
\label{sec:sec4-ft}
We also investigate whether task-specific fine-tuned models can provide evaluations that better align with human judgments than their general-purpose counterparts. To this end, we conduct fine-tuning experiments on Llama-3.1-8B\footnote{\scriptsize{\href{https://hf.co/meta-llama/Llama-3.1-8B}{https://hf.co/meta-llama/Llama-3.1-8B}}}.

\section{Results}

\subsection{Experimental Setup}

This section presents the results of our experiments following the methodology outlined in Section~\ref{sec:eval_methodology}. We first compare the performance of previously adopted evaluation metrics used in the target shared task~\cite{Dementieva2024OverviewOT}, which introduced the multilingual \textdetox\ dataset. We then analyze the performance of our proposed metrics and examine how LLMs perform as automatic evaluators. Finally, we report results from LLM fine-tuning experiments designed to establish comprehensive benchmarks across multiple aspects of the text detoxification task. The results obtained on the DialogueEvaluation-2022 dataset in Russian~\cite{Dementieva2022RUSSE2022} are reported in Appendix \ref{sec:appendix_russe2022}.

To ensure a fair comparison between LLM-as-a-judge and the fine-tuning experiments, we split the \textdetox\ dataset by dividing the 20 participating systems into training and test sets, where outputs from 14 systems are used for training, while outputs from the remaining 6 systems are reserved for testing. This participant-based split ensures that fine-tuned models are evaluated on outputs from previously unseen systems. All results reported for both LLM-as-a-judge and fine-tuning experiments are computed on this held-out test set.

To assess the quality and reliability of different evaluation approaches, we measure the agreement between automatic metrics and human judgments using Spearman's rank correlation coefficient ($\rho$). This non-parametric measure is particularly well-suited for our study as it captures monotonic relationships without assuming linearity, making it robust to the varying scales and distributions inherent in different metric families. For each evaluation dimension---\textit{fluency}, \textit{content similarity}, and \textit{toxicity}---we compute correlations between metric scores and human annotations across all system outputs. We report correlations separately for each language to account for linguistic variation and to identify cross-lingual patterns in metric behavior.

\begin{figure}[!h]
\centering
\tiny
\includegraphics[width=0.49\textwidth]{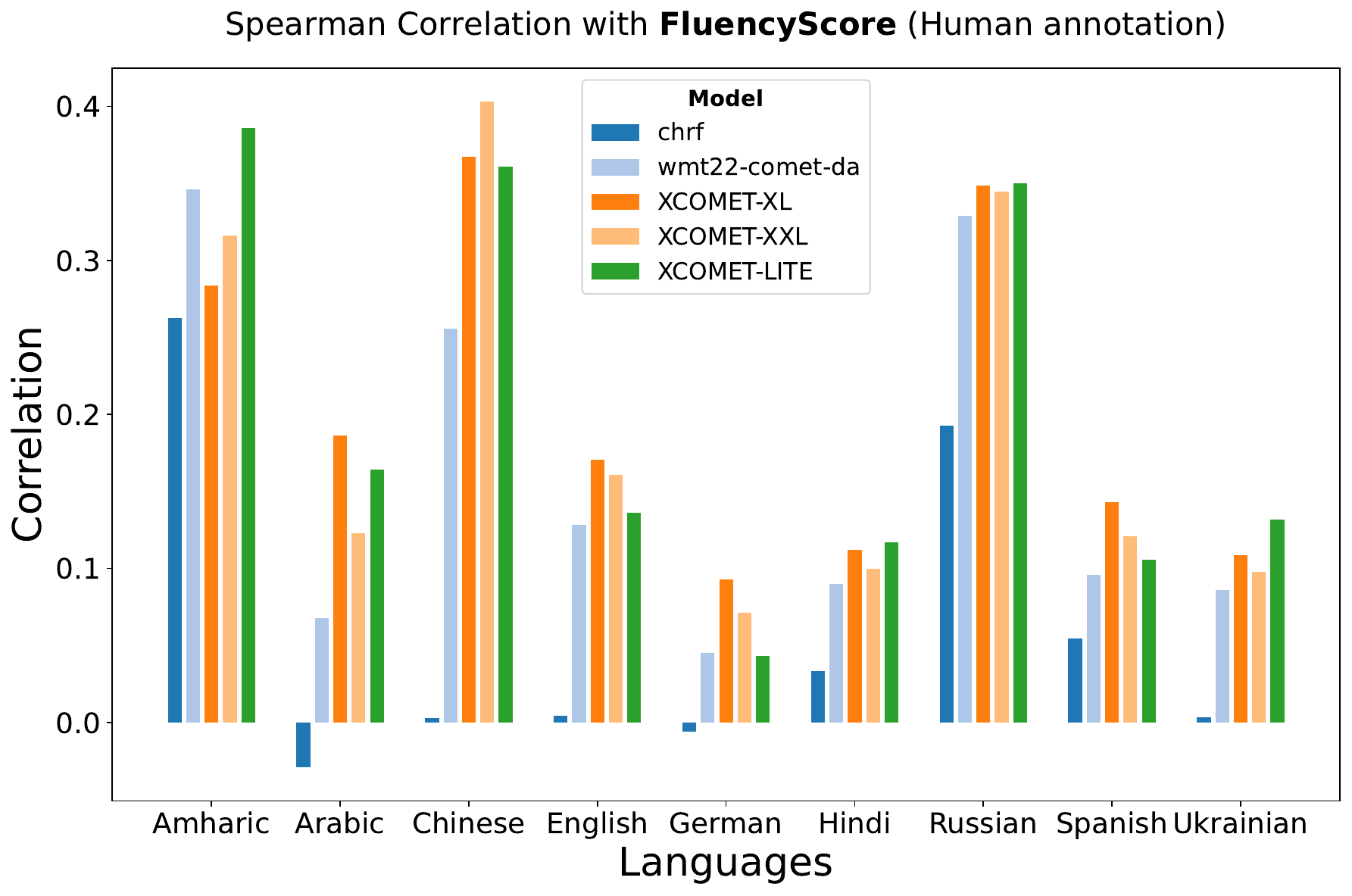}
\caption{\textdetox: Correlation of fluency measurement approaches with human-annotated fluency scores.}
\label{fig:fluency}
\end{figure}

\subsection{Fluency Evaluation Results}

Figure~\ref{fig:fluency} presents Spearman correlations between automatic fluency metrics and human judgments across all languages and models. ChrF consistently underperforms, showing near-zero correlations in Arabic, German, Chinese, English, and Ukrainian, which can be attributed to its reliance on character $n$-gram overlap with references, penalizing semantically equivalent paraphrases that use different word choices or phrasing, a common requirement in detoxification where toxic phrases must be substantially rewritten. In contrast, COMET-based models demonstrate positive correlations across languages, with \textbf{XCOMET-XXL} and \textbf{XCOMET-lite} achieving the strongest performance in Amharic, Chinese, and Russian. The magnitude of correlations varies considerably, ranging from weak in German and Hindi to moderate in Amharic and Russian, suggesting that fluency evaluation difficulty is language-dependent. This variation may stem from differences in training data availability for these languages in XCOMET's multilingual pre-training, as well as varying degrees of lexical diversity in the reference translations across languages. Notably, a quantized model \textbf{XCOMET-lite} maintains competitive performance with \textbf{XCOMET-XXL}, making it an optimal choice for future detoxification competitions and production systems that require reliable evaluation without extensive computational resources.

\subsection{Content Similarity Evaluation Results}

\begin{figure}[!h]
\centering
\tiny
\includegraphics[width=0.49\textwidth]{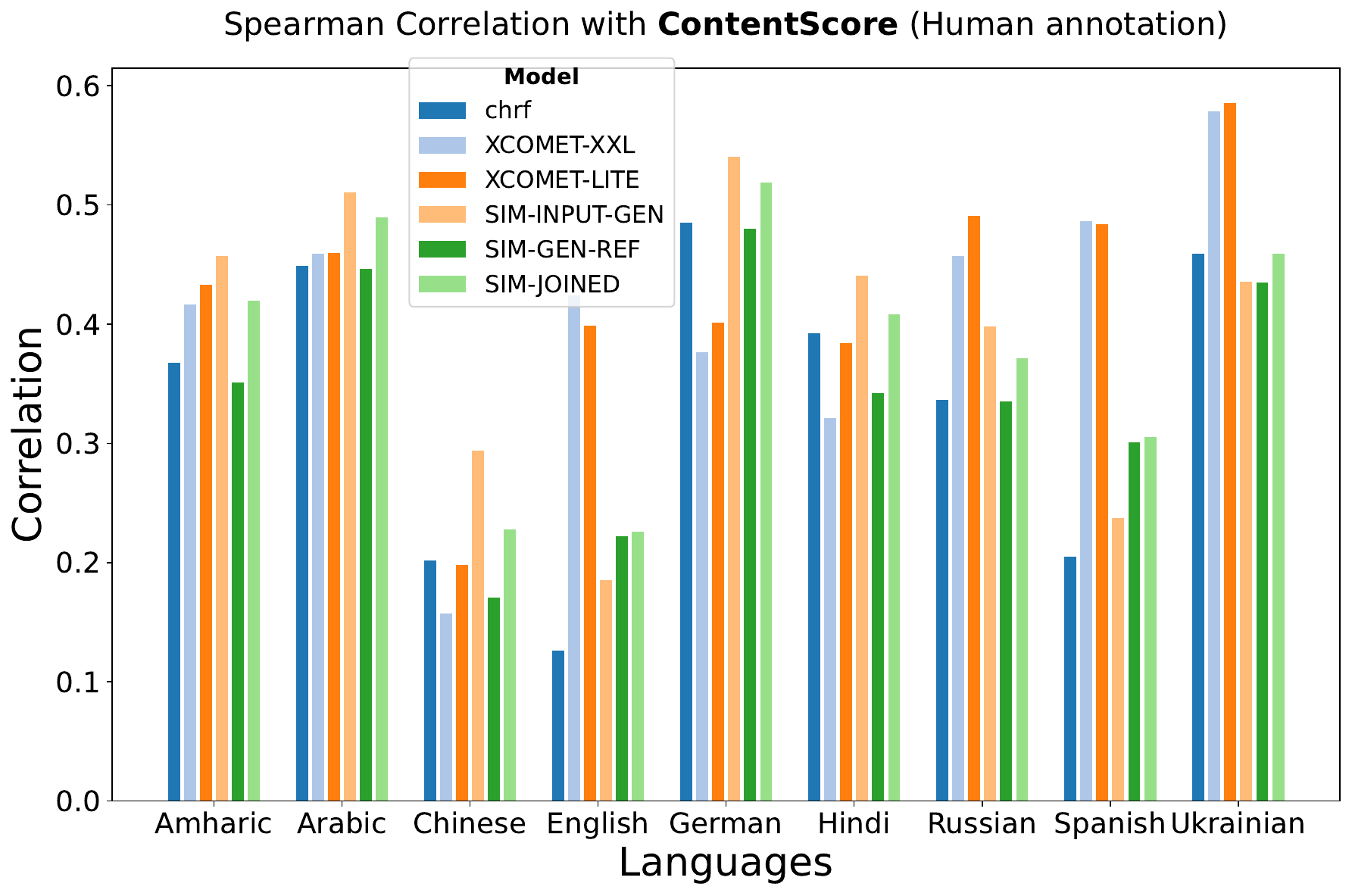}
\caption{\textdetox: Correlation of content similarity measurement approaches with human-annotated content preservation scores.}
\label{fig:content}
\end{figure}
Figure~\ref{fig:content} presents Spearman correlations between automatic content similarity metrics and human judgments across all languages. We evaluate three embedding-based configurations: (i) \textbf{SIM-INPUT-GEN}, cosine similarity between the toxic input and generated output, serving as the baseline approach from prior work~\cite{Dementieva2024OverviewOT}; (ii) \textbf{SIM-GEN-REF}, similarity between the generated output and human reference; and (iii) \textbf{SIM-JOINED}, our proposed weighted combination defined in Equation~\ref{eq:1}, where $w_{i,g} = 0.4$ and $w_{g,r} = 0.6$. We assign higher weight to the generated-reference similarity because human references represent high-quality detoxification examples that balance content preservation with appropriate rewording, while input-output similarity may be misleading when toxic phrasing requires substantial changes.

Surprisingly, the results reveal that the baseline \textbf{SIM-INPUT-GEN} achieves the highest correlations with human judgments in five languages (Amharic, Arabic, Chinese, German, Hindi). This unexpected finding suggests that human annotators may prioritize preserving the original text’s meaning over adhering to the reference paraphrasing style when assessing content preservation. However, such a tendency can be detrimental to annotation consistency, as it may favor outputs that closely mirror the source text and penalize significant paraphrases that achieve equivalent meaning through different lexical or syntactic choices. In languages where detoxification requires substantial rewording to remove offensive content (Arabic, German), measuring similarity only to toxic inputs provides limited signal about whether appropriate semantic transfer has occurred. Among neural models, \textbf{XCOMET-LITE} and \textbf{XCOMET-XXL} demonstrate strong and consistent performance across all languages. Notably, ChrF shows surprisingly competitive performance in Arabic, German, and Ukrainian, suggesting that character $n$-gram overlap can approximate content preservation when detoxification strategies involve minor lexical substitutions.

According to our results, we recommend prioritizing COMET-based models for evaluating content similarity in detoxification tasks. Unlike embedding-based metrics that capture only surface-level similarity to the input, XCOMET models jointly consider the relationships among the input, output, and reference texts. This joint modeling yields more stable and robust evaluations that are more resistant to superficial lexical changes. Moreover, the direct consideration of reference texts in XCOMET aligns more closely with the detoxification objective, producing fluent, neutral outputs that preserve meaning without reproducing toxic expressions.

\begin{figure}[!h]
\centering
\tiny
\includegraphics[width=0.5\textwidth]{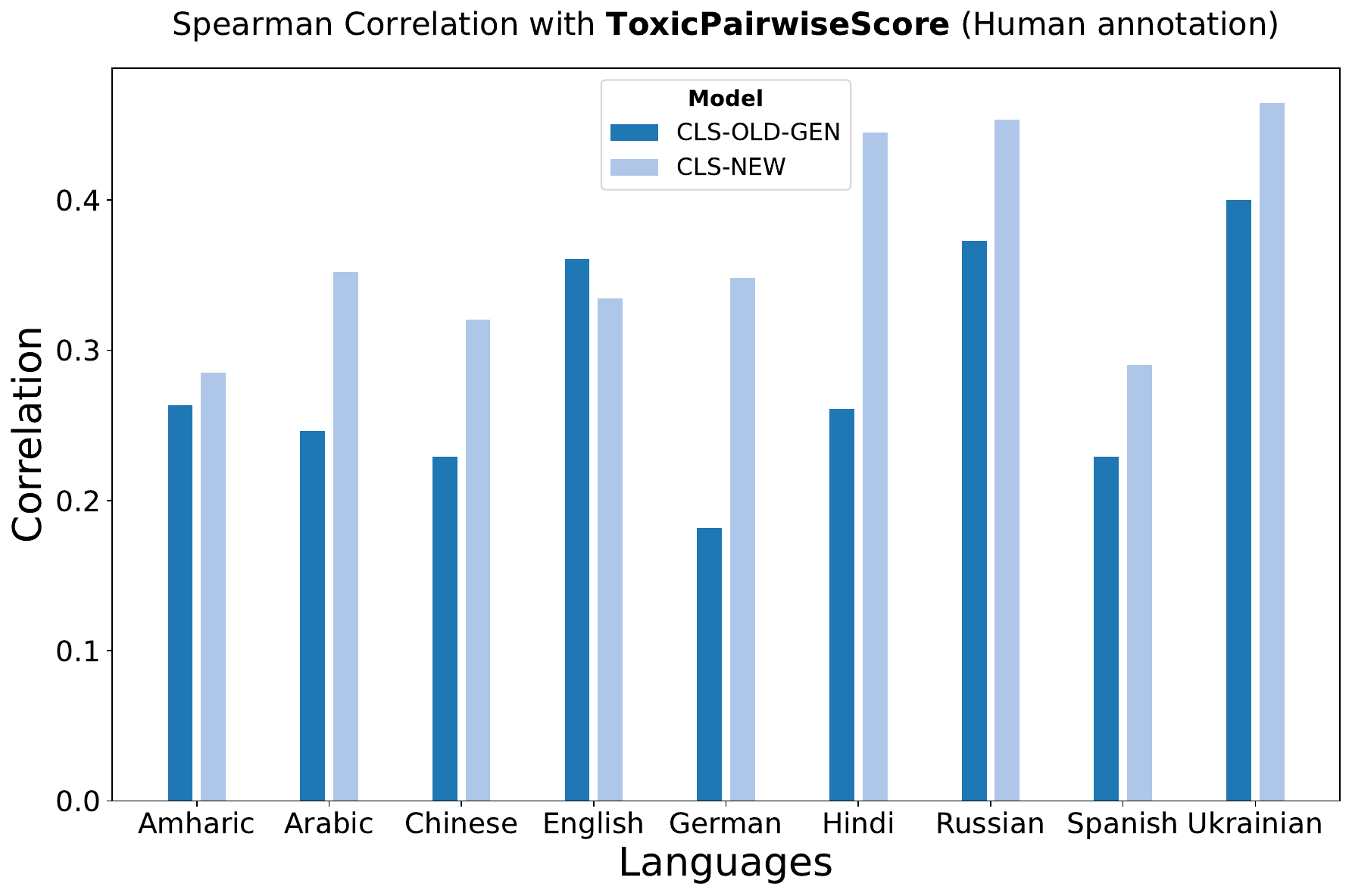}
\caption{\textdetox: Correlation of toxicity measurement approaches with target pairwise toxic human annotated scores.}
\label{fig:toxic}
\end{figure}

\subsection{Toxicity Evaluation Results}

As described in Section~\ref{sec:sta_desc}, the effectiveness of style transfer in detoxification is typically assessed using a pretrained toxicity classifier, which estimates the probability that a generated text belongs to the \textit{non-toxic} class.
Here we aim to compare two approaches:  
(i) \textbf{CLS-OLD-GEN}, which measures the predicted probability of the generated text being non-toxic; and  
(ii) \textbf{CLS-NEW}, a joint probability–based approach introduced in Section~\ref{sec:sta_desc}, which integrates toxicity signals from the input, generated output, and reference texts.  
Both approaches employ the same model\footnote{\scriptsize{\href{https://hf.co/textdetox/xlmr-large-toxicity-classifier}{https://hf.co/textdetox/xlmr-large-toxicity-classifier}}} as prior work.

Figure~\ref{fig:toxic} presents the correlation of these approaches with human-annotated toxicity scores. \textbf{CLS-NEW} achieves the highest correlation across all languages except English, where \textbf{CLS-OLD-GEN} performs slightly better.
Strong cross-lingual performance of \textbf{CLS-NEW} highlights the advantage of incorporating contextual information from input–output–reference triplets. By jointly considering them, \textbf{CLS-NEW} provides more stable toxicity estimates, reducing sensitivity to minor lexical changes and enabling fairer, context-sensitive evaluation in the detoxification task.

\subsection{Evaluation of Joined Metric (J)}
\begin{figure}[!h]
\centering
\tiny
\includegraphics[width=0.5\textwidth]{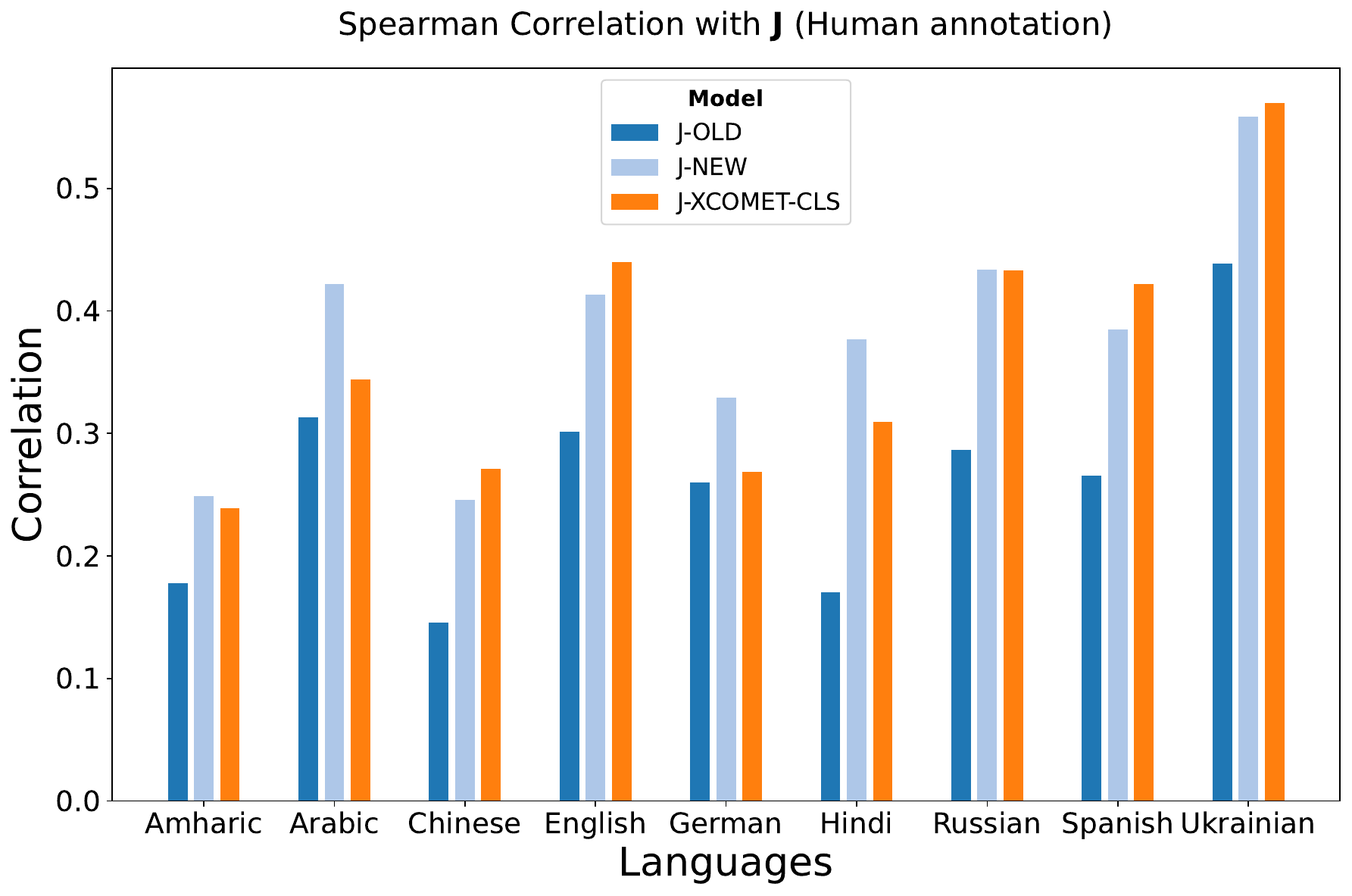}
\caption{\textdetox: Correlation final scores with target joined scores from human annotation.}
\label{fig:J}
\end{figure}
Following the original formulation of the joint detoxification score presented in Equation~\ref{eq:J}, we evaluate how well different metric combinations correlate with human judgments. Figure~\ref{fig:J} presents results for three composite evaluation approaches: (i) \textbf{J-OLD}, the baseline approach combining ChrF, SIM-INPUT-GEN, and CLS-OLD-GEN; (ii) \textbf{J-NEW}, our proposed combination of XCOMET-LITE, SIM-JOINED, and CLS-NEW; and (iii) \textbf{J-XCOMET-CLS}, a simplified variant using only XCOMET-LITE and CLS-NEW. The rationale for considering J-XCOMET-CLS is to leverage XCOMET-LITE's ability to assess both fluency and content similarity from input–output–reference triplets, eliminating the need for a separate content similarity metric.

The results show that \textbf{J-OLD} achieves the lowest correlations across all languages, highlighting the limitations of the baseline approach. Our proposed \textbf{J-NEW} method demonstrates the highest correlations in 5 languages (Amharic, Arabic, German, Hindi, Russian). Interestingly, \textbf{J-XCOMET-CLS} performs best in 4 languages (Chinese, English, Spanish, Ukrainian), slightly outperforming \textbf{J-NEW}, but showing lower correlations in Arabic, German, and Hindi. These findings suggest that XCOMET-LITE can effectively capture both fluency and content similarity in the text detoxification task.

\begin{figure*}[!t]
\centering
\tiny
\includegraphics[width=0.9\textwidth]{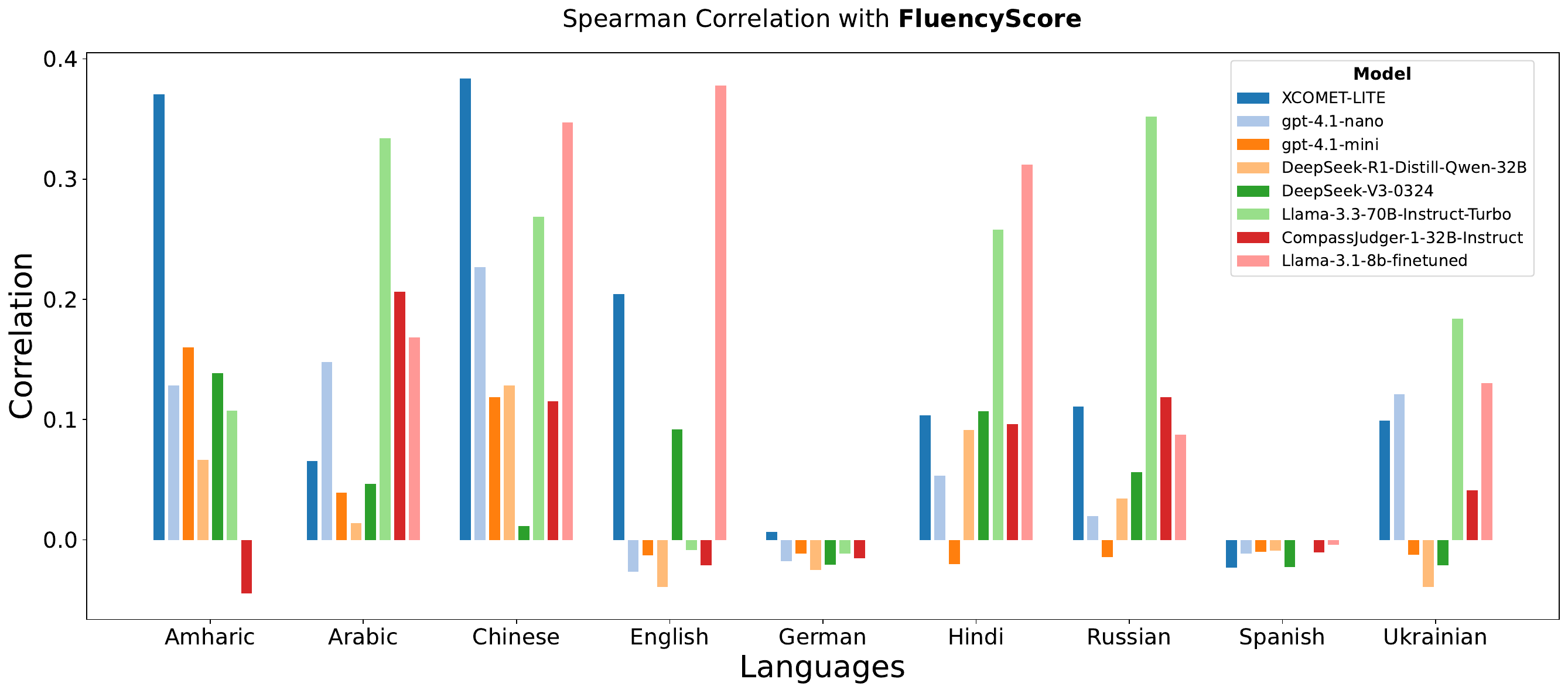}
\caption{\textdetox: Comparison between \textbf{XCOMET-LITE} and different LLMs on the fluency scores from human annotation.}
\label{fig:fluency2}
\end{figure*}

\begin{figure*}[!t]
\centering
\tiny
\includegraphics[width=0.9\textwidth]{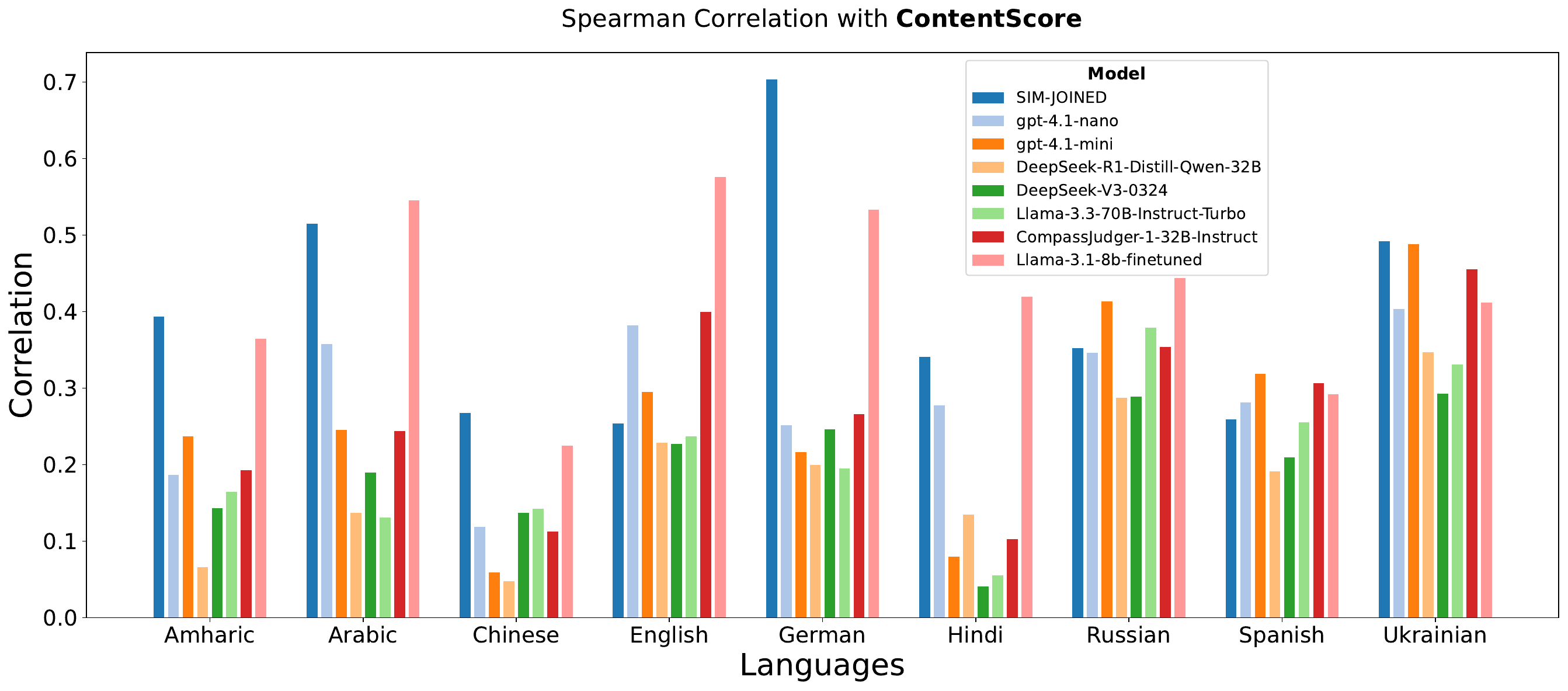}
\caption{\textdetox: Comparison between \textbf{SIM-JOINED} and different LLMs on the content similarity scores from human annotation.}
\label{fig:content2}
\end{figure*}

\begin{figure*}[!t]
\centering
\tiny
\includegraphics[width=0.9\textwidth]{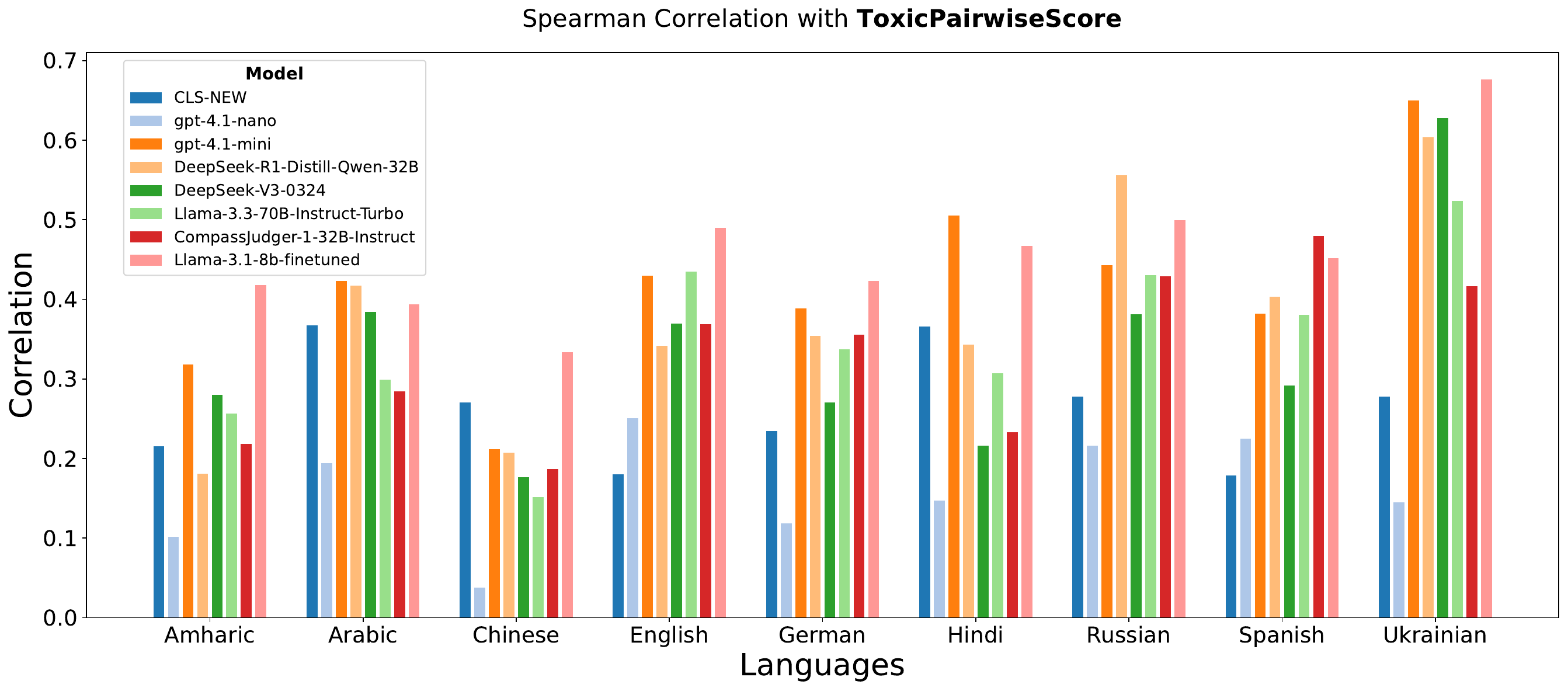}
\caption{\textdetox: Comparison between \textbf{CLS-NEW} and different LLMs on the toxicity classification scores from human annotation.}
\label{fig:toxic2}
\end{figure*}

\subsection{Evaluation using LLMs as Judges}
\label{sec:sec5-llmjudge}
In addition to aforementioned methods, we also explore the use of LLMs as evaluators. We consider a diverse set of models: DeepSeek-R1-Distill-Qwen-32B\footnote{\scriptsize{\href{https://hf.co/deepseek-ai/DeepSeek-R1-Distill-Qwen-32B}{https://hf.co/deepseek-ai/DeepSeek-R1-Distill-Qwen-32B}}}, DeepSeek-V3-0324\footnote{\scriptsize{\href{https://hf.co/deepseek-ai/DeepSeek-V3-0324}{https://hf.co/deepseek-ai/DeepSeek-V3-0324}}}, LLaMA 3.3-70B-Instruct\footnote{\scriptsize{\href{https://hf.co/meta-llama/Llama-3.3-70B-Instruct}{https://hf.co/meta-llama/Llama-3.3-70B-Instruct}}}, CompassJudger-1-32B-Instruct~\cite{cao2024compass}, and GPT-4.1 variants (\textit{nano} and \textit{mini}).  
These models represent a diverse range of architectures and scales, allowing us to examine whether larger or more instruction-tuned models offer improved alignment with human judgments across detoxification evaluation dimensions. All final prompts are provided in Appendix \ref{sec:app_prompts}.

\paragraph{Fluency}
Figure~\ref{fig:fluency2} depicts correlation results with human annotations, comparing XCOMET-LITE against six LLM-as-a-judge approaches (Llama-3.1-8b-finetuned is considered in the next subsection). XCOMET-LITE significantly outperforms LLM baselines only in Amharic and Chinese. In contrast, LLaMA 3.3-70B-Instruct-Turbo achieves the highest correlations in Arabic, Hindi, Russian, and Ukrainian, substantially surpassing other models.
\paragraph{Content similarity}
Figure~\ref{fig:content2} compares the proposed SIM-JOINED metric with six LLM-as-a-judge approaches. SIM-JOINED achieves the highest correlations in Amharic, Arabic, Chinese, German, Hindi, and Ukrainian, outperforming LLM-based evaluation in most cases. This is in contrast to fluency evaluation, where XCOMET-LITE performed worse than LLMs across most languages.
This difference can be explained  by the nature of the tasks: content similarity is a clear semantic matching problem, where embedding-based metrics perform well by measuring the proximity of text representations.
In contrast, fluency requires assessing grammaticality, naturalness, and stylistic appropriateness factors that XCOMET-LITE struggles to capture as effectively as LLMs in this detoxification task.
\paragraph{Toxicity}
In Figure~\ref{fig:toxic2} the CLS-NEW model outperforms all LLMs only in Chinese and shows poor correlations in other languages. Notably, GPT-4.1-mini demonstrates strong performance across all languages; DeepSeek-R1-Distill-Qwen-32B achieves the highest correlations in Russian, significantly outperforming all other models, and demonstrates one of the best performance in Arabic and Spanish.

\subsection{Fine-tuning Experiments}
\label{sec:sec5-ft}
We perform fine-tuning of Llama-3.1-8B using the Low-Rank Adaptation (LoRA)~\cite{hu2022lora} method to efficiently adapt the base models while minimizing computational overhead. The main model weights are loaded in 4-bit quantized format, enabling faster training and reduced memory usage without significant loss in performance. The LoRA configuration utilizes the following hyperparameters: rank $r=8$, $\alpha=16$, and a dropout rate of $0.1$, with adaptation applied to all linear layers.
The fine-tuning process was conducted for $2$ epochs over the training data. For each task (i.e. content similarity, style transfer, or fluence) we fine-tuned standalone LORA. The resulting tuned LLMs are available in our HuggingFace repository.\footnote{\scriptsize{\href{https://hf.co/textdetox/Llama-pairwise-toxicity-evaluator}{https://hf.co/textdetox/Llama-pairwise-toxicity-evaluator}}}$^,$\footnote{\scriptsize{\href{https://hf.co/textdetox/Llama-pairwise-content-evaluator}{https://hf.co/textdetox/Llama-pairwise-content-evaluator}}}

Figures~\ref{fig:fluency2}, \ref{fig:content2}, and~\ref{fig:toxic2} present the fine-tuning results for Llama-3.1-8B across all evaluation dimensions. For fluency evaluation, the fine-tuned model achieves the highest correlations in English and demonstrates competitive performance in Chinese, while showing notably lower correlations in other languages. This pattern likely reflects the composition of Llama's pretraining data, where English is substantially more represented, enabling better adaptation to fluency assessment tasks in that language. For content similarity evaluation, the fine-tuned model demonstrates strong and consistent performance across nearly all languages, achieving the highest or near-highest correlations except in Spanish and Ukrainian. Finally, for toxicity evaluation, the fine-tuned model shows stable performance across all languages, achieving the highest correlations in Amharic, Chinese, English, German, and Ukrainian. 
In summary, the findings suggest that each evaluation task benefits differently from fine-tuning, with fluency being most sensitive to language coverage, while content and toxicity assessments exhibit stronger cross-lingual robustness.

\section{Conclusion}
We presented the first large-scale multilingual evaluation of text detoxification on the \textdetox\ dataset, covering all core dimensions: fluency, content similarity, and toxicity reduction. To improve evaluation, we introduced XCOMET-based models for fluency, a triplet-based SIM-JOINED metric for content similarity that leverages input–output–reference texts, along with CLS-NEW, a refined toxicity metric that jointly considers probability distributions also across all text triplets.
We thoroughly analyzed different evaluation metrics and LLM configurations, including LLM-as-a-judge experiments and task-specific fine-tuning, to identify approaches that align most closely with human judgments across diverse languages.
We hope the benchmark results, improved metrics, and insights into evaluation approaches presented in this study facilitate future research on multilingual text detoxification and related style transfer tasks.

%=================================================================================================================================================================================================================================================================

\section{Limitations}

The experiments in this paper are limited by the number of languages considered. Currently, our conclusions are based solely on experiments involving 9 languages: Amharic, Arabic, Chinese, English, German, Hindi, Russian, Spanish, and Ukrainian. As a result, our findings may be biased toward these languages due to the lack of available data for others. The generalization of our proposed metrics to additional languages, particularly low-resource languages outside these families, remains an open question requiring further investigation.

Our evaluation relies on the \textdetox\ and DialogueEvaluation-2022 datasets, which contain outputs from 20 and 15 detoxification systems, respectively. While these datasets represent diverse modeling approaches, the limited number of systems and the specific characteristics of their outputs may influence metric correlations. Future work should validate our findings on larger and more diverse collections of detoxification systems across additional domains.

Proposed XCOMET-based fluency metrics and fine-tuned LLM evaluators depend on the quality and coverage of their pre-training data. Languages with limited representation in these training corpora may receive less reliable evaluations. The observed performance variations across languages particularly the weaker correlations in German and Hindi suggest that data availability during model pre-training directly impacts evaluation quality.

\section{Ethical Considerations}
Our work on evaluating text detoxification is motivated by the goal of fostering safer, more respectful online communication, rather than restricting freedom of expression. By focusing on developing better metrics for assessing detoxification quality, we aim to support the responsible deployment of detoxification models—ensuring that such systems are evaluated not only for effectiveness, but also for fairness, transparency, and contextual nuance.

We believe that detoxification tools, when applied, should serve as suggestions rather than enforcement mechanisms. Ideally, these tools would be integrated with user-centric interfaces that allow individuals to make informed decisions about their language use, with the final choice remaining in the hands of the user.

\section{Acknowledgments}\label{sec:ack}

The work of Alexander Panchenko was supported by the RSF № 25-71-30008 ``Laboratory for reliable, adaptive, and trustworthy Artificial Intelligence''.

\section{Bibliographical References}\label{sec:reference}

\bibliographystyle{lrec2026-natbib}
\bibliography{lrec2026-example}

@article{DBLP:journals/corr/abs-2502-15022,
  author       = {Amalie Brogaard Pauli and
                  Isabelle Augenstein and
                  Ira Assent},
  title        = {A Meta-Evaluation of Style and Attribute Transfer Metrics},
  journal      = {CoRR},
  volume       = {abs/2502.15022},
  year         = {2025},
  url          = {https://doi.org/10.48550/arXiv.2502.15022},
  doi          = {10.48550/ARXIV.2502.15022},
  eprinttype    = {arXiv},
  eprint       = {2502.15022},
  bibsource    = {dblp computer science bibliography, https://dblp.org}
}

@inproceedings{Popovic2015chrFCN,
  title={chrF: character n-gram F-score for automatic MT evaluation},
  author={Maja Popovic},
  booktitle={WMT@EMNLP},
  year={2015},
  url={https://api.semanticscholar.org/CorpusID:15349458}
}

@inproceedings{Lin2004ROUGEAP,
  title={ROUGE: A Package for Automatic Evaluation of Summaries},
  author={Chin-Yew Lin},
  booktitle={Annual Meeting of the Association for Computational Linguistics},
  year={2004},
  url={https://api.semanticscholar.org/CorpusID:964287}
}

@inproceedings{Banerjee2005METEORAA,
    title = "{METEOR}: An Automatic Metric for {MT} Evaluation with Improved Correlation with Human Judgments",
    author = "Banerjee, Satanjeev  and
      Lavie, Alon",
    editor = "Goldstein, Jade  and
      Lavie, Alon  and
      Lin, Chin-Yew  and
      Voss, Clare",
    booktitle = "Proceedings of the {ACL} Workshop on Intrinsic and Extrinsic Evaluation Measures for Machine Translation and/or Summarization",
    month = jun,
    year = "2005",
    address = "Ann Arbor, Michigan",
    publisher = "Association for Computational Linguistics",
    url = "https://aclanthology.org/W05-0909/",
    pages = "65--72"
}

@article{Zhang2019BERTScoreET,
  title={BERTScore: Evaluating Text Generation with BERT},
  author={Tianyi Zhang and Varsha Kishore and Felix Wu and Kilian Q. Weinberger and Yoav Artzi},
  journal={ArXiv},
  year={2019},
  volume={abs/1904.09675},
  url={https://api.semanticscholar.org/CorpusID:127986044}
}

@inproceedings{Bavaresco2024LLMsIO,
    title = "{LLM}s instead of Human Judges? A Large Scale Empirical Study across 20 {NLP} Evaluation Tasks",
    author = "Bavaresco, Anna  and
      Bernardi, Raffaella  and
      Bertolazzi, Leonardo  and
      Elliott, Desmond  and
      Fern{\'a}ndez, Raquel  and
      Gatt, Albert  and
      Ghaleb, Esam  and
      Giulianelli, Mario  and
      Hanna, Michael  and
      Koller, Alexander  and
      Martins, Andre  and
      Mondorf, Philipp  and
      Neplenbroek, Vera  and
      Pezzelle, Sandro  and
      Plank, Barbara  and
      Schlangen, David  and
      Suglia, Alessandro  and
      Surikuchi, Aditya K  and
      Takmaz, Ece  and
      Testoni, Alberto",
    editor = "Che, Wanxiang  and
      Nabende, Joyce  and
      Shutova, Ekaterina  and
      Pilehvar, Mohammad Taher",
    booktitle = "Proceedings of the 63rd Annual Meeting of the Association for Computational Linguistics (Volume 2: Short Papers)",
    month = jul,
    year = "2025",
    address = "Vienna, Austria",
    publisher = "Association for Computational Linguistics",
    url = "https://aclanthology.org/2025.acl-short.20/",
    doi = "10.18653/v1/2025.acl-short.20",
    pages = "238--255",
    ISBN = "979-8-89176-252-7"
}

@string{acl = {Association for Computational Linguistics}}

@string{anth = {https://aclanthology.org/}}

@string{NAACL:2021:main = {Proceedings of the 2021 Conference of the North American Chapter of the Association for Computational Linguistics: Human Language Technologies}}

@string{ACL:2018:2 = {Proceedings of the 56th Annual Meeting of the Association for Computational Linguistics (Volume 2: Short Papers)}}

@string{EMNLP:2021:main = {Proceedings of the 2021 Conference on Empirical Methods in Natural Language Processing}}

@string{ACL:2023:short = {Proceedings of the 61st Annual Meeting of the Association for Computational Linguistics (Volume 2: Short Papers)}}

@string{ACL:2022:long = {Proceedings of the 60th Annual Meeting of the Association for Computational Linguistics (Volume 1: Long Papers)}}

@string{ICON:2023:1 = {Proceedings of the 20th International Conference on Natural Language Processing (ICON)}}

@string{IJCNLP:2023:main = {Proceedings of the 13th International Joint Conference on Natural Language Processing and the 3rd Conference of the Asia-Pacific Chapter of the Association for Computational Linguistics (Volume 1: Long Papers)}}

@string{NAACL:2024:short = {Proceedings of the 2024 Conference of the North American Chapter of the Association for Computational Linguistics: Human Language Technologies (Volume 2: Short Papers)}}

@string{FINDINGS:2024:naacl = {Findings of the Association for Computational Linguistics: NAACL 2024}}

@string{COLING:2025:main = {Proceedings of the 31st International Conference on Computational Linguistics}}

@inproceedings{Dementieva2024OverviewOT,
  author       = {Daryna Dementieva and
                  Daniil Moskovskiy and
                  Nikolay Babakov and
                  Abinew Ali Ayele and
                  Naquee Rizwan and
                  Florian Schneider and
                  Xintong Wang and
                  Seid Muhie Yimam and
                  Dmitry Ustalov and
                  Elisei Stakovskii and
                  Alisa Smirnova and
                  Ashraf Elnagar and
                  Animesh Mukherjee and
                  Alexander Panchenko},
  editor       = {Guglielmo Faggioli and
                  Nicola Ferro and
                  Petra Galusc{\'{a}}kov{\'{a}} and
                  Alba Garc{\'{\i}}a Seco de Herrera},
  title        = {Overview of the Multilingual Text Detoxification Task at {PAN} 2024},
  booktitle    = {Working Notes of the Conference and Labs of the Evaluation Forum {(CLEF}
                  2024), Grenoble, France, 9-12 September, 2024},
  series       = {{CEUR} Workshop Proceedings},
  volume       = {3740},
  pages        = {2432--2461},
  publisher    = {CEUR-WS.org},
  year         = {2024},
  url          = {https://ceur-ws.org/Vol-3740/paper-223.pdf},
  bibsource    = {dblp computer science bibliography, https://dblp.org}
}

@article{Dementieva2022RUSSE2022,
	title        = {{RUSSE-2022: Findings of the First Russian Detoxification Shared Task Based on Parallel Corpora}},
	author       = {Daryna Dementieva and Varvara Logacheva and Irina Nikishina and Alena Fenogenova and David Dale and I. Krotova and Nikita Semenov and Tatiana Shavrina and Alexander Panchenko},
	year         = 2022,
	journal      = {COMPUTATIONAL LINGUISTICS AND INTELLECTUAL TECHNOLOGIES},
	url          = {https://api.semanticscholar.org/CorpusID:253169495}
}

@inproceedings{Feng2020LanguageagnosticBS,
    title = "Language-agnostic {BERT} Sentence Embedding",
    author = "Feng, Fangxiaoyu  and
      Yang, Yinfei  and
      Cer, Daniel  and
      Arivazhagan, Naveen  and
      Wang, Wei",
    editor = "Muresan, Smaranda  and
      Nakov, Preslav  and
      Villavicencio, Aline",
    booktitle = "Proceedings of the 60th Annual Meeting of the Association for Computational Linguistics (Volume 1: Long Papers)",
    month = may,
    year = "2022",
    address = "Dublin, Ireland",
    publisher = "Association for Computational Linguistics",
    url = "https://aclanthology.org/2022.acl-long.62/",
    doi = "10.18653/v1/2022.acl-long.62",
    pages = "878--891"
}

@inproceedings{Conneau2019UnsupervisedCR,
    title = "Unsupervised Cross-lingual Representation Learning at Scale",
    author = "Conneau, Alexis  and
      Khandelwal, Kartikay  and
      Goyal, Naman  and
      Chaudhary, Vishrav  and
      Wenzek, Guillaume  and
      Guzm{\'a}n, Francisco  and
      Grave, Edouard  and
      Ott, Myle  and
      Zettlemoyer, Luke  and
      Stoyanov, Veselin",
    editor = "Jurafsky, Dan  and
      Chai, Joyce  and
      Schluter, Natalie  and
      Tetreault, Joel",
    booktitle = "Proceedings of the 58th Annual Meeting of the Association for Computational Linguistics",
    month = jul,
    year = "2020",
    address = "Online",
    publisher = "Association for Computational Linguistics",
    url = "https://aclanthology.org/2020.acl-main.747/",
    doi = "10.18653/v1/2020.acl-main.747",
    pages = "8440--8451"
}

@article{Guerreiro2023xcometTM,
  title={xcomet: Transparent Machine Translation Evaluation through Fine-grained Error Detection},
  author={Nuno M. Guerreiro and Ricardo Rei and Daan van Stigt and Lu{\'i}sa Coheur and Pierre Colombo and Andr{\'e} Martins},
  journal={Transactions of the Association for Computational Linguistics},
  year={2023},
  volume={12},
  pages={979-995},
  url={https://api.semanticscholar.org/CorpusID:264146484}
}

@article{Rei2020COMETAN,
  title={COMET: A Neural Framework for MT Evaluation},
  author={Ricardo Rei and Craig Alan Stewart and Ana C. Farinha and Alon Lavie},
  journal={ArXiv},
  year={2020},
  volume={abs/2009.09025},
  url={https://api.semanticscholar.org/CorpusID:221819581}
}

@inproceedings{Rei2022COMET22U2,
  title={COMET-22: Unbabel-IST 2022 Submission for the Metrics Shared Task},
  author={Ricardo Rei and Jos{\'e} G. C. de Souza and Duarte M. Alves and Chrysoula Zerva and Ana C. Farinha and T. Glushkova and Alon Lavie and Lu{\'i}sa Coheur and Andr{\'e} F. T. Martins},
  booktitle={Conference on Machine Translation},
  year={2022},
  url={https://api.semanticscholar.org/CorpusID:256461051}
}

@Article{Carlson2018,
  author        = "Carlson, Keith and Riddell, Allen and Rockmore, Daniel",
  title         = "Evaluating prose style transfer with the Bible",
  journal       = "Royal Society Open Science",
  volume        = "5",
  month         = "October",
  year          = "2018",
  OPTurl           = "https://doi.org/10.1098/rsos.171920",
  url={https://arxiv.org/abs/1711.04731}
}

@inproceedings{pryzant2020automatically,
  title={Automatically neutralizing subjective bias in text},
  author={Pryzant, Reid and Martinez, Richard Diehl and Dass, Nathan and Kurohashi, Sadao and Jurafsky, Dan and Yang, Diyi},
  booktitle={Proceedings of the aaai conference on artificial intelligence},
  volume={34},
  pages={480--489},
  year={2020}
}

@inproceedings{smurfcat_at_pan,
  author       = {Elisei Rykov and
                  Konstantin Zaytsev and
                  Ivan Anisimov and
                  Alexandr Voronin},
  editor       = {Guglielmo Faggioli and
                  Nicola Ferro and
                  Petra Galusc{\'{a}}kov{\'{a}} and
                  Alba Garc{\'{\i}}a Seco de Herrera},
  title        = {SmurfCat at {PAN} 2024 TextDetox: Alignment of Multilingual Transformers
                  for Text Detoxification},
  booktitle    = {Working Notes of the Conference and Labs of the Evaluation Forum {(CLEF}
                  2024), Grenoble, France, 9-12 September, 2024},
  series       = {{CEUR} Workshop Proceedings},
  volume       = {3740},
  pages        = {2866--2871},
  publisher    = {CEUR-WS.org},
  year         = {2024},
  url          = {https://ceur-ws.org/Vol-3740/paper-276.pdf},
  bibsource    = {dblp computer science bibliography, https://dblp.org}
}

@article{DBLP:journals/corr/abs-2406-18403,
  author       = {Anna Bavaresco and
                  Raffaella Bernardi and
                  Leonardo Bertolazzi and
                  Desmond Elliott and
                  Raquel Fern{\'{a}}ndez and
                  Albert Gatt and
                  Esam Ghaleb and
                  Mario Giulianelli and
                  Michael Hanna and
                  Alexander Koller and
                  Andr{\'{e}} F. T. Martins and
                  Philipp Mondorf and
                  Vera Neplenbroek and
                  Sandro Pezzelle and
                  Barbara Plank and
                  David Schlangen and
                  Alessandro Suglia and
                  Aditya K. Surikuchi and
                  Ece Takmaz and
                  Alberto Testoni},
  title        = {LLMs instead of Human Judges? {A} Large Scale Empirical Study across
                  20 {NLP} Evaluation Tasks},
  journal      = {CoRR},
  volume       = {abs/2406.18403},
  year         = {2024},
  url          = {https://doi.org/10.48550/arXiv.2406.18403},
  doi          = {10.48550/ARXIV.2406.18403},
  eprinttype    = {arXiv},
  eprint       = {2406.18403},
  bibsource    = {dblp computer science bibliography, https://dblp.org}
}

@article{DBLP:journals/corr/abs-2412-05579,
  author       = {Haitao Li and
                  Qian Dong and
                  Junjie Chen and
                  Huixue Su and
                  Yujia Zhou and
                  Qingyao Ai and
                  Ziyi Ye and
                  Yiqun Liu},
  title        = {LLMs-as-Judges: {A} Comprehensive Survey on LLM-based Evaluation Methods},
  journal      = {CoRR},
  volume       = {abs/2412.05579},
  year         = {2024},
  url          = {https://doi.org/10.48550/arXiv.2412.05579},
  doi          = {10.48550/ARXIV.2412.05579},
  eprinttype    = {arXiv},
  eprint       = {2412.05579},
  bibsource    = {dblp computer science bibliography, https://dblp.org}
}

@inproceedings{DBLP:conf/clef/PengHZYLLGCLT24,
  author       = {Jiangao Peng and
                  Zhongyuan Han and
                  Huan Zhang and
                  Jingyan Ye and
                  Chang Liu and
                  Biao Liu and
                  Mingcan Guo and
                  Haoyang Chen and
                  Zijie Lin and
                  Yujiao Tang},
  editor       = {Guglielmo Faggioli and
                  Nicola Ferro and
                  Petra Galusc{\'{a}}kov{\'{a}} and
                  Alba Garc{\'{\i}}a Seco de Herrera},
  title        = {A Multilingual Text Detoxification Method Based on Few-shot Learning
                  and {CO-STAR} Framework},
  booktitle    = {Working Notes of the Conference and Labs of the Evaluation Forum {(CLEF}
                  2024), Grenoble, France, 9-12 September, 2024},
  series       = {{CEUR} Workshop Proceedings},
  volume       = {3740},
  pages        = {2829--2837},
  publisher    = {CEUR-WS.org},
  year         = {2024},
  url          = {https://ceur-ws.org/Vol-3740/paper-271.pdf},
  bibsource    = {dblp computer science bibliography, https://dblp.org}
}

@inproceedings{DBLP:conf/clef/LuoLW24,
  author       = {Zhongyu Luo and
                  Man Luo and
                  Aiguo Wang},
  editor       = {Guglielmo Faggioli and
                  Nicola Ferro and
                  Petra Galusc{\'{a}}kov{\'{a}} and
                  Alba Garc{\'{\i}}a Seco de Herrera},
  title        = {Multilingual Text Detoxification Using Google Cloud Translation and
                  Post-Processing},
  booktitle    = {Working Notes of the Conference and Labs of the Evaluation Forum {(CLEF}
                  2024), Grenoble, France, 9-12 September, 2024},
  series       = {{CEUR} Workshop Proceedings},
  volume       = {3740},
  pages        = {2769--2774},
  publisher    = {CEUR-WS.org},
  year         = {2024},
  url          = {https://ceur-ws.org/Vol-3740/paper-263.pdf},
  bibsource    = {dblp computer science bibliography, https://dblp.org}
}

@inproceedings{DBLP:conf/clef/Protasov24,
  author       = {Vitaly Protasov},
  editor       = {Guglielmo Faggioli and
                  Nicola Ferro and
                  Petra Galusc{\'{a}}kov{\'{a}} and
                  Alba Garc{\'{\i}}a Seco de Herrera},
  title        = {{PAN} 2024 Multilingual TextDetox: Exploring Cross-lingual Transfer
                  Using Large Language Models},
  booktitle    = {Working Notes of the Conference and Labs of the Evaluation Forum {(CLEF}
                  2024), Grenoble, France, 9-12 September, 2024},
  series       = {{CEUR} Workshop Proceedings},
  volume       = {3740},
  pages        = {2852--2857},
  publisher    = {CEUR-WS.org},
  year         = {2024},
  url          = {https://ceur-ws.org/Vol-3740/paper-274.pdf},
  bibsource    = {dblp computer science bibliography, https://dblp.org}
}

@string{NAACL:2018:1 = {Proceedings of the 2018 Conference of the North {A}merican Chapter of the Association for Computational Linguistics: Human Language Technologies, Volume 1 (Long Papers)}}

@inproceedings{rao-tetreault-2018-dear,title = "Dear Sir or Madam, May {I} Introduce the {GYAFC} Dataset: Corpus, Benchmarks and Metrics for Formality Style Transfer",author = "Rao, Sudha and Tetreault, Joel",editor = "Walker, Marilyn and Ji, Heng and Stent, Amanda",booktitle = NAACL:2018:1,month = jun,year = "2018",address = "New Orleans, Louisiana",
publisher = acl,
url = anth # {N18-1012/},
doi = "10.18653/v1/N18-1012",pages = "129--140"
}

@inproceedings{li-etal-2018-delete,title = "Delete, Retrieve, Generate: a Simple Approach to Sentiment and Style Transfer",author = "Li, Juncen and Jia, Robin and He, He and Liang, Percy",editor = "Walker, Marilyn and Ji, Heng and Stent, Amanda",booktitle = NAACL:2018:1,month = jun,year = "2018",address = "New Orleans, Louisiana",publisher = acl,url = anth # {N18-1169/},doi = "10.18653/v1/N18-1169",pages = "1865--1874"
}

@inproceedings{briakou-etal-2021-ola,title = "Ol{\'a}, Bonjour, Salve! {XFORMAL}: A Benchmark for Multilingual Formality Style Transfer",author = "Briakou, Eleftheria and Lu, Di and Zhang, Ke and Tetreault, Joel",editor = "Toutanova, Kristina and Rumshisky, Anna and Zettlemoyer, Luke and Hakkani-Tur, Dilek and Beltagy, Iz and Bethard, Steven and Cotterell, Ryan and Chakraborty, Tanmoy and Zhou, Yichao",booktitle = NAACL:2021:main,month = jun,year = "2021",address = "Online",publisher = acl,url = anth # {2021.naacl-main.256/},doi = "10.18653/v1/2021.naacl-main.256",pages = "3199--3216"
}

@inproceedings{nogueira-dos-santos-etal-2018-fighting,title = "Fighting Offensive Language on Social Media with Unsupervised Text Style Transfer",author = "Nogueira dos Santos, Cicero and Melnyk, Igor and Padhi, Inkit",editor = "Gurevych, Iryna and Miyao, Yusuke",booktitle = ACL:2018:2,month = jul,year = "2018",address = "Melbourne, Australia",publisher = acl,url = anth # {P18-2031/},doi = "10.18653/v1/P18-2031",pages = "189--194"
}

@inproceedings{dale-etal-2021-text,title = "Text Detoxification using Large Pre-trained Neural Models",author = "Dale, David and Voronov, Anton and Dementieva, Daryna and Logacheva, Varvara and Kozlova, Olga and Semenov, Nikita and Panchenko, Alexander",editor = "Moens, Marie-Francine and Huang, Xuanjing and Specia, Lucia and Yih, Scott Wen-tau",booktitle = EMNLP:2021:main,month = nov,year = "2021",address = "Online and Punta Cana, Dominican Republic",publisher = acl,url = anth # {2021.emnlp-main.629/},doi = "10.18653/v1/2021.emnlp-main.629",pages = "7979--7996"
}

@inproceedings{hallinan-etal-2023-detoxifying,title = "Detoxifying Text with {M}a{RC}o: Controllable Revision with Experts and Anti-Experts",author = "Hallinan, Skyler and Liu, Alisa and Choi, Yejin and Sap, Maarten",editor = "Rogers, Anna and Boyd-Graber, Jordan and Okazaki, Naoaki",booktitle = ACL:2023:short,month = jul,year = "2023",address = "Toronto, Canada",publisher = acl,url = anth # {2023.acl-short.21/},doi = "10.18653/v1/2023.acl-short.21",pages = "228--242"
}

@inproceedings{logacheva-etal-2022-paradetox,title = "{P}ara{D}etox: Detoxification with Parallel Data",author = "Logacheva, Varvara and Dementieva, Daryna and Ustyantsev, Sergey and Moskovskiy, Daniil and Dale, David and Krotova, Irina and Semenov, Nikita and Panchenko, Alexander",editor = "Muresan, Smaranda and Nakov, Preslav and Villavicencio, Aline",booktitle = ACL:2022:long,month = may,year = "2022",address = "Dublin, Ireland",publisher = acl,url = anth # {2022.acl-long.469/},doi = "10.18653/v1/2022.acl-long.469",pages = "6804--6818"
}

@inproceedings{sourabrata-etal-2023-text,title = "Text Detoxification as Style Transfer in {E}nglish and {H}indi",author = "Mukherjee, Sourabrata and Bansal, Akanksha and Kr. Ojha, Atul and P. McCrae, John and Dusek, Ondrej",editor = "D. Pawar, Jyoti and Lalitha Devi, Sobha",booktitle = ICON:2023:1,month = dec,year = "2023",address = "Goa University, Goa, India",publisher = "NLP Association of India (NLPAI)",url = anth # {2023.icon-1.13/},pages = "133--144"
}

@inproceedings{dementieva-etal-2023-exploring,title = "Exploring Methods for Cross-lingual Text Style Transfer: The Case of Text Detoxification",author = "Dementieva, Daryna and Moskovskiy, Daniil and Dale, David and Panchenko, Alexander",editor = "Park, Jong C. and Arase, Yuki and Hu, Baotian and Lu, Wei and Wijaya, Derry and Purwarianti, Ayu and Krisnadhi, Adila Alfa",booktitle = IJCNLP:2023:main,month = nov,year = "2023",address = "Nusa Dua, Bali",publisher = acl,url = anth # {2023.ijcnlp-main.70/},doi = "10.18653/v1/2023.ijcnlp-main.70",pages = "1083--1101"
}

@inproceedings{jones-etal-2024-multi,title = "A Multi-Aspect Framework for Counter Narrative Evaluation using Large Language Models",author = "Jones, Jaylen and Mo, Lingbo and Fosler-Lussier, Eric and Sun, Huan",editor = "Duh, Kevin and Gomez, Helena and Bethard, Steven",booktitle = NAACL:2024:short,month = jun,year = "2024",address = "Mexico City, Mexico",publisher = acl,url = anth # {2024.naacl-short.14/},doi = "10.18653/v1/2024.naacl-short.14",pages = "147--168"
}

@inproceedings{bonaldi-etal-2024-nlp,title = "{NLP} for Counterspeech against Hate: A Survey and How-To Guide",author = "Bonaldi, Helena and Chung, Yi-Ling and Abercrombie, Gavin and Guerini, Marco",editor = "Duh, Kevin and Gomez, Helena and Bethard, Steven",booktitle = FINDINGS:2024:naacl,month = jun,year = "2024",address = "Mexico City, Mexico",publisher = acl,url = anth # {2024.findings-naacl.221/},doi = "10.18653/v1/2024.findings-naacl.221",pages = "3480--3499"
}

@inproceedings{lemesle-etal-2025-paraphrase,title = "Paraphrase Generation Evaluation Powered by an {LLM}: A Semantic Metric, Not a Lexical One",author = "Lemesle, Quentin and Chevelu, Jonathan and Martin, Philippe and Lolive, Damien and Delhay, Arnaud and Barbot, Nelly",editor = "Rambow, Owen and Wanner, Leo and Apidianaki, Marianna and Al-Khalifa, Hend and Eugenio, Barbara Di and Schockaert, Steven",booktitle = COLING:2025:main,month = jan,year = "2025",address = "Abu Dhabi, UAE",publisher = acl,url = anth # {2025.coling-main.538/},pages = "8057--8087"
}

@inproceedings{larionov-etal-2024-xcomet,
    title = "x{COMET}-lite: Bridging the Gap Between Efficiency and Quality in Learned {MT} Evaluation Metrics",
    author = "Larionov, Daniil  and
      Seleznyov, Mikhail  and
      Viskov, Vasiliy  and
      Panchenko, Alexander  and
      Eger, Steffen",
    editor = "Al-Onaizan, Yaser  and
      Bansal, Mohit  and
      Chen, Yun-Nung",
    booktitle = "Proceedings of the 2024 Conference on Empirical Methods in Natural Language Processing",
    month = nov,
    year = "2024",
    address = "Miami, Florida, USA",
    publisher = "Association for Computational Linguistics",
    url = "https://aclanthology.org/2024.emnlp-main.1223",
    pages = "21934--21949",
}

@inproceedings{shen-etal-2022-evaluation,
    title = "On the Evaluation Metrics for Paraphrase Generation",
    author = "Shen, Lingfeng  and
      Liu, Lemao  and
      Jiang, Haiyun  and
      Shi, Shuming",
    editor = "Goldberg, Yoav  and
      Kozareva, Zornitsa  and
      Zhang, Yue",
    booktitle = "Proceedings of the 2022 Conference on Empirical Methods in Natural Language Processing",
    month = dec,
    year = "2022",
    address = "Abu Dhabi, United Arab Emirates",
    publisher = "Association for Computational Linguistics",
    url = "https://aclanthology.org/2022.emnlp-main.208/",
    doi = "10.18653/v1/2022.emnlp-main.208",
    pages = "3178--3190"
}

@article{cao2024compass,
  author       = {Maosong Cao and
                  Alexander Lam and
                  Haodong Duan and
                  Hongwei Liu and
                  Songyang Zhang and
                  Kai Chen},
  title        = {CompassJudger-1: All-in-one Judge Model Helps Model Evaluation and
                  Evolution},
  journal      = {CoRR},
  volume       = {abs/2410.16256},
  year         = {2024},
  url          = {https://doi.org/10.48550/arXiv.2410.16256},
  doi          = {10.48550/ARXIV.2410.16256},
  eprinttype    = {arXiv},
  eprint       = {2410.16256},
  bibsource    = {dblp computer science bibliography, https://dblp.org}
}

@inproceedings{hu2022lora,
  author       = {Edward J. Hu and
                  Yelong Shen and
                  Phillip Wallis and
                  Zeyuan Allen{-}Zhu and
                  Yuanzhi Li and
                  Shean Wang and
                  Lu Wang and
                  Weizhu Chen},
  title        = {LoRA: Low-Rank Adaptation of Large Language Models},
  booktitle    = {The Tenth International Conference on Learning Representations, {ICLR}
                  2022, Virtual Event, April 25-29, 2022},
  publisher    = {OpenReview.net},
  year         = {2022},
  url          = {https://openreview.net/forum?id=nZeVKeeFYf9},
  bibsource    = {dblp computer science bibliography, https://dblp.org}
}

% \section{Language Resource References}
% \label{lr:ref}
% \bibliographystylelanguageresource{lrec2026-natbib}
% \bibliographylanguageresource{languageresource}
\onecolumn
\appendix

\section{Licensing of Resources}
\label{sec:app_licenses}

Below is an overview of the licenses associated with each resource used in this work (Table~\ref{tab:overview-license}).

\begin{table}[ht!]
\centering
\footnotesize
\begin{tabular}{p{4cm}p{2.5cm}p{8cm}}
\toprule
Resource & License & Homepage  \\ 
\midrule
TextDetoxEval-Style & OpenRail++ & \href{https://hf.co/datasets/textdetox/detoxification_pairwise_style_evaluation}{https://hf.co/datasets/textdetox/detoxification\_pairwise\_\newline style\_evaluation}\\
TextDetoxEval-Content & OpenRail++ & \href{https://hf.co/datasets/textdetox/detoxification_pairwise_content_evaluation}{https://hf.co/datasets/textdetox/detoxification\_pairwise\_\newline content\_evaluation} \\
TextDetoxEval-RU & OpenRail++ & \href{https://hf.co/datasets/textdetox/humaneval_textdetox_ru}{https://hf.co/datasets/textdetox/humaneval\_textdetox\_ru} \\
Llama-pairwise-toxicity-evaluator & OpenRail++ & \href{https://hf.co/textdetox/Llama-pairwise-toxicity-evaluator}{https://hf.co/textdetox/Llama-pairwise-toxicity-evaluator} \\
Llama-pairwise-content-evaluator & OpenRail++ & \href{https://hf.co/textdetox/Llama-pairwise-content-evaluator}{https://hf.co/textdetox/Llama-pairwise-content-evaluator} \\
\midrule
LLaMa3 & llama3 & \href{https://hf.co/meta-llama}{https://hf.co/meta-llama} \\
DeepSeek & MIT & \href{https://hf.co/collections/deepseek-ai/deepseek-r1-678e1e131c0169c0bc89728d}{https://hf.co/collections/deepseek-ai/deepseek-r1-678e1e131c0169c0bc89728d}\\

LaBSE & Apache 2.0 & \href{https://hf.co/sentence-transformers/LaBSE}{https://hf.co/sentence-transformers/LaBSE} \\

xlmr-large-toxicity-classifier & Open RAIL++ & \href{https://hf.co/textdetox/xlmr-large-toxicity-classifier}{https://hf.co/textdetox/xlmr-large-toxicity-classifier} \\
XCOMET-lite & - & \href{https://hf.co/myyycroft/XCOMET-lite}{https://hf.co/myyycroft/XCOMET-lite} \\
wmt22-comet-da & Apache 2.0 & \href{https://hf.co/Unbabel/wmt22-comet-da}{https://hf.co/Unbabel/wmt22-comet-da} \\
XCOMET-XL & CC-BY-NC-SA-4.0 & \href{https://hf.co/Unbabel/XCOMET-XL}{https://hf.co/Unbabel/XCOMET-XL} \\
XCOMET-XXL & CC-BY-NC-SA-4.0 & \href{https://hf.co/Unbabel/XCOMET-XXL}{https://hf.co/Unbabel/XCOMET-XXL} \\
\bottomrule
\end{tabular}
\caption{Overview of the licenses associated with each resource.}
\label{tab:overview-license}
\end{table}

% DeepSeek-R1-Distill-Qwen-32B\footnote{\scriptsize{\href{https://hf.co/deepseek-ai/DeepSeek-R1-Distill-Qwen-32B}{https://hf.co/deepseek-ai/DeepSeek-R1-Distill-Qwen-32B}}}, DeepSeek-V3-0324\footnote{\scriptsize{\href{https://hf.co/deepseek-ai/DeepSeek-V3-0324}{https://hf.co/deepseek-ai/DeepSeek-V3-0324}}}, LLaMA 3.3-70B-Instruct\footnote{\scriptsize{\href{https://hf.co/meta-llama/Llama-3.3-70B-Instruct}{https://hf.co/meta-llama/Llama-3.3-70B-Instruct}}}

The licenses associated with the models and datasets used in this study are consistent with the intended purpose of conducting academic research aimed at advancing various NLP applications for positive impact.

\section{LLMs Prompts}
\label{sec:app_prompts}

Here, we provide exact prompts used for LLMs prompting.

\begin{tcolorbox}[breakable, colback=black!5!white,        % Very light gray background
  colframe=gray!10!black,      % Light black/gray frame color
  width=\linewidth,       % Set the width of the box
  boxrule=0.4mm,          % Thickness of the border
  arc=2mm,                % Rounded corners
  outer arc=2mm,          % Outer rounded corners
  boxsep=2.5mm,             % Space between box and text
  title=\texttt{Fluency prompt}]
  
You are given with the text and your task is to evaluate the naturalness or fluency of the text.

Options:

yes (meaning that the text is natural and fluent)

partially (meaning that the text is partially natural and fluent)

no (meaning that the text is not natural and fluent at all)

Examples:

text: "I am going to the store to buy some groceries."

Answer: yes

text: "The cat sat on the mat dog argh adfsdf"

Answer: partially

text: "teh dsfsdf ats 1324@3 adf"

Answer: no

text: 
Answer:
\end{tcolorbox}

\begin{tcolorbox}[breakable, colback=black!5!white,        % Very light gray background
  colframe=gray!10!black,      % Light black/gray frame color
  width=\linewidth,       % Set the width of the box
  boxrule=0.4mm,          % Thickness of the border
  arc=2mm,                % Rounded corners
  outer arc=2mm,          % Outer rounded corners
  boxsep=2.5mm,             % Space between box and text
  title=\texttt{Content similarity prompt}]
  
You are given with two texts. Your task is to compare the meaning of them and decide whether they mean the same or not.
\\\\
Options:
\\
- yes (both texts have the same meaning)
\\
- no (there is significant difference in meaning between the two texts)
\\\\
Examples:

text1: "Why do you behave like an asshole?"
\\
text2: "Why do you act so bad?"
\\
Answer: yes
\\\\
text1: "This political pary is a gang of idiots"

text2: "They just dont know their stuff"

Answer: no
\\\\
text1: 

text2:

Answer:
\end{tcolorbox}

\begin{tcolorbox}[breakable, colback=black!5!white,        % Very light gray background
  colframe=gray!10!black,      % Light black/gray frame color
  width=\linewidth,       % Set the width of the box
  boxrule=0.4mm,          % Thickness of the border
  arc=2mm,                % Rounded corners
  outer arc=2mm,          % Outer rounded corners
  boxsep=2.5mm,             % Space between box and text
  title=\texttt{Style transfer evaluation prompt}]
  
You are given with two texts. Your task is to compare which text is more toxic.

Options:

- text1 (text1 is more toxic than text2)

- text2 (text2 is more toxic than text1)

- None (both texts have equal toxicity - both of them are either toxic or non-toxic at the same time)

Examples:

text1: "I hate you, you are the worst person ever."

text2: "I love you, you are the best person ever."

Answer: text1

text1: "You are so smart, I can't believe you did that."

text2: "You are so stupid, I can't believe you did that."

Answer: text2

text1: "I think you are a great person."

text2: "I think you are a very good man"

Answer: none

You must return ONLY one of the options. Do not add any explanations or additional information.

text1: 
text2: 
Answer:
\end{tcolorbox}

\section{DialogueEvaluation-2022 Dataset}
\label{sec:appendix_russe2022}

\subsection{Dataset description}

The DialogueEvaluation-2022 detoxification dataset was released as part of the first competition on Russian-language text detoxification. The dataset comprises toxic sentences sourced from Russian social media platforms such as Odnoklassniki, Pikabu, and Twitter.
Data collection followed a three-stage crowdsourcing pipeline using Yandex.Toloka. First, annotators were asked to rewrite toxic sentences into fluent, non-toxic paraphrases while preserving the original meaning. Second, independent annotators verified that the paraphrases retained the same semantic content as the originals. Third, additional annotators checked whether the paraphrases were indeed non-toxic. Only paraphrases that passed both the content and toxicity checks (with $\geq$ 90\% agreement) were included in the dataset.

Annotations were performed according to strict guidelines that defined toxicity and provided examples. Human evaluation focused on three components: \textit{fluency} (grammatical correctness and readability), \textit{content preservation} (semantic similarity with the original), and \textit{style transfer quality} (removal of toxicity).

\subsection{Results}

\begin{figure}[!h]
\centering
\tiny
\includegraphics[width=1\textwidth]{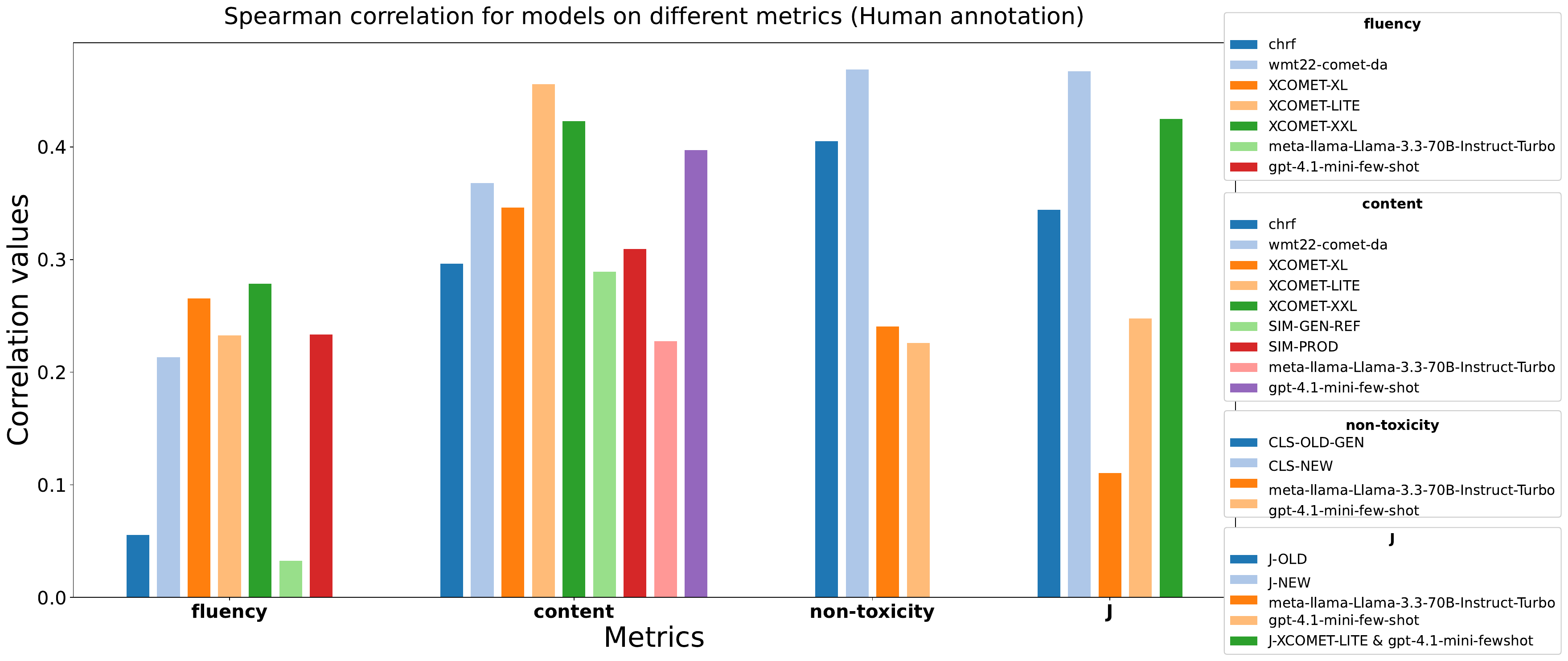}
\caption{DialogueEvaluation-2022 results across different models.}
\label{fig:content-dialogue2022}
\end{figure}

As we can see, for the \textbf{fluency} estimation, \textbf{XCOMET-XXL} still shows the best results, while \textbf{XCOMET-XL} performs slightly worse. At the same time, \textbf{gpt-4.1-mini-few-shot} demonstrates comparable results to \textbf{XCOMET-LITE}, though both fall short of the two aforementioned models.

Regarding \textbf{content similarity}, \textbf{XCOMET-LITE} achieves the best performance, while \textbf{gpt-4.1-mini-few-shot} yields results comparable to \textbf{XCOMET-XXL}.

For distinguishing between toxic and non-toxic texts, our new \textbf{CLS-PROD} model attains the highest scores, surpassing both the previous \textbf{CLS-OLD-GEN} model and the two considered LLMs. This is a surprising observation, as it directly contradicts the results from Figure~\ref{fig:toxic2} for the \textdetox\ dataset, where our models performed significantly worse than nearly all LLMs for Russian.
Finally, the joint evaluation scores are highest for the proposed model (\textbf{J-PROD}), outperforming all other models.

\end{document}